\pdfoutput=1
\documentclass[11pt]{article}

\usepackage[margin=1in]{geometry}
\usepackage{graphicx}
\usepackage{booktabs}
\usepackage{amsmath}
\usepackage[authoryear]{natbib}
\usepackage{authblk}
\usepackage{caption}
\usepackage[hidelinks]{hyperref}
\usepackage{microtype}

\title{\bfseries Do vision-language models search like humans?\\
\large Reasoning tokens as a reaction-time analog in classic visual-search paradigms}

\author[1]{Farahnaz Wick}
\affil[1]{Independent Researcher \\ \texttt{farahnaz@gmail.com}}
\date{June 2026}

\begin{document}
\maketitle

\begin{abstract}
\noindent
Visual search has been one of the most productive paradigms in the study of visual
attention: the way reaction time scales with the number of items distinguishes
parallel, ``pop-out'' search from serial, attention-demanding search. I ask
whether vision-language models (VLMs) exhibit the same behavioral signatures. I
adapt four classic paradigms: feature versus conjunction search, spatial-configuration
(\,T-vs-L\,) search, enumeration, and the tilted/vertical search
asymmetry; and present them to current frontier and mid-tier models. Because a
single model call has no reaction time, I use the number of reasoning
(``thinking'') tokens a model spends per trial as a within-model analog of search
effort, and I compare against a large public human benchmark
\citep{wolfe2010}. The models reproduce several human signatures: feature search
costs flat effort while conjunction effort climbs with set size; frontier models
hold accuracy where mid-tier models collapse to chance; and a resolution control
shows the conjunction cost is genuine search rather than difficulty resolving
small shapes. They also diverge from humans in informative ways. The
target-present effort slope exceeds the target-absent slope, reversing the human
ordering; enumeration remains accurate where humans would lose count; and a
reasoning model with adaptive deliberation declines to deliberate on detection
tasks altogether, so that a single search expresses itself as an effort gradient
in one model and as an accuracy cliff in another. I argue that psychophysical
paradigms, applied behaviorally, are a sharp and inexpensive probe of machine
visual cognition, and that the points of divergence are as informative as the
points of agreement.
\end{abstract}

\section{Introduction}

Vision-language models now describe, caption, and reason about images
fluently, which raises a question that fluency alone cannot answer: when a model
locates an object in a cluttered scene, is it doing anything like what a person
does, or does it merely produce human-like outputs through a non-human process?

Visual search offers a clean way to study this question. A century of work has
established that the function relating response time to the number of items in a
display---the search slope---carries diagnostic information about the underlying
mechanism \citep{treisman1980,wolfe2021}. A target that differs from its
distractors in a single basic feature is found in roughly constant time
regardless of how many distractors are present; the slope is flat, the hallmark
of a parallel, preattentive process. A target defined only by the conjunction of
features must be found by inspecting items more or less serially, and each added
distractor adds time; the slope is positive. Flat versus sloped is a fingerprint
of two different mechanisms, and the paradigm has been extended to enumeration,
to search asymmetries, and to spatial-configuration search, each with its own
characteristic signature.

This paper asks whether VLMs show these signatures. The central methodological
obstacle is that reaction time, the dependent measure of the entire human
literature, does not exist for a single forward pass of a model. I address this
by exploiting a feature of contemporary reasoning models: when allowed to
deliberate, they emit a variable number of intermediate ``reasoning'' or
``thinking'' tokens before committing to an answer \citep{wei2022}. I treat the
number of reasoning tokens spent on a trial as a within-model analog of search
effort. If a model performs anything like serial search, it should spend more
reasoning on conjunction-type trials as the display grows and stay flat on
feature trials, mirroring the human reaction-time slope.

Using this measure I run four experiments. The first
(Section~\ref{sec:fc}) is the canonical feature-versus-conjunction comparison,
run across six model configurations and benchmarked against approximately
$75{,}910$ human trials. A resolution control (Section~\ref{sec:blur})
asks whether the conjunction cost reflects search or merely the difficulty of
resolving small letters. The remaining three experiments remove the color cue
that the first study still relied on: a spatial-configuration T-vs-L search
(Section~\ref{sec:tvl}), an enumeration task (Section~\ref{sec:enum}), and the
classic tilted-versus-vertical search asymmetry (Section~\ref{sec:asym}). My
contributions are: (i) a behavioral, psychophysics-style protocol for probing
VLM visual search that requires no access to model internals; (ii) the use of
reasoning-token counts as a within-model effort measure benchmarked against human
reaction-time patterns; and (iii) a catalog of agreements and, more
interestingly, divergences between machine and human search.

\section{Background and related work}
\label{sec:background}

\paragraph{Feature integration and guided search.}
\citet{treisman1980} proposed that simple features such as color and orientation
are registered preattentively and in parallel across the visual field, whereas
binding features into objects requires focused attention deployed serially. The
empirical signature is the contrast between flat feature-search slopes and
positive conjunction-search slopes. Guided Search \citep{wolfe1989,wolfe2021}
refined this picture: attention is not deployed at random but is \emph{guided}
toward likely targets by a small set of preattentive feature maps, color
foremost among them, combined into a priority map. Guidance is why conjunction
search, though serial, is far more efficient than exhaustive inspection. It is
also why the design of a search display matters: a conjunction target that shares
its color with half the distractors can still be approached by restricting
attention to the correct color subset.

\paragraph{Target-absent search and asymmetries.}
Two further regularities are central here. First, in inefficient search the
target-absent slope is typically steeper than the target-present slope, often by
a factor near two, because confirming that nothing is present requires inspecting
more of the display than stumbling onto a target that is present
\citep{wolfe2010,palmer2011}. Second, search can be asymmetric: a target bearing
a feature its distractors lack (a tilted bar among vertical bars) is found
efficiently, whereas the same discrimination in the opposite direction (a
vertical bar among tilted bars) is inefficient \citep{treisman1988}. Asymmetries
are a strong test because the two directions use identical elements; any
difference must come from how the system represents the feature.

\paragraph{Enumeration.}
Reporting how many targets are present dissociates into two regimes: rapid,
accurate \emph{subitizing} of up to about four items, and slow, error-prone
\emph{counting} beyond that range, which is thought to require serial deployment
of attention \citep{trick1994}. Enumeration thus offers a second axis of serial
load, orthogonal to detection.

\paragraph{The serial signature.}
A recurring theme across these paradigms is that the diagnostic quantity is not
the absolute response time but the \emph{slope}: a linear rise in time with the
relevant load parameter (set size or number of items) is the canonical behavioral
signature of a serial, attention-demanding operation
\citep{treisman1980,trick1994}. \citet{ullman1984} formalized this family of
operations as \emph{visual routines}, sequences of elementary operations applied
to a visual representation. I adopt the same logic here: my claims rest on the
shape of effort as a function of load, not on its absolute magnitude.

\paragraph{Reasoning tokens and effort.}
Contemporary language and vision-language models can be prompted or configured to
produce an explicit chain of intermediate tokens before answering
\citep{wei2022}. The length of this chain varies with problem difficulty and is
exposed in API usage metadata. I use it as an effort proxy. The analogy to
reaction time is imperfect---tokens are discrete, generated autoregressively, and
shaped by training rather than by oculomotor and attentional dynamics---but it
shares the key property I need: it is a graded, per-trial cost that the model
allocates itself, without instruction, as a function of difficulty.

\paragraph{VLMs and visual perception.}
A growing literature documents that VLMs, despite strong benchmark scores, fail on
visual tasks that are trivial for humans. \citet{rahmanzadehgervi2024} show that
state-of-the-art models perform poorly on elementary perceptual judgments such as
whether two lines intersect or how many shapes overlap, and \citet{fu2024} find
that perception-demanding tasks which ``resist mediation through natural
language'' remain near chance. \citet{campbell2024} attribute a large class of
these failures to the \emph{binding problem}: the difficulty of using a shared
representational substrate to keep distinct objects and their features apart.
\citet{budny2025} sharpen this into a \emph{serial processing} account, showing
across geometric reasoning, enumeration, and mental rotation that VLM accuracy
declines precisely where human reaction time (their proxy for serial load)
increases.

My work is complementary in measure and method. Prior work indexes difficulty
with human reaction time and reads out VLM \emph{accuracy}; I instead read a
within-model \emph{effort} signal directly from reasoning-token counts and ask
whether its shape, across the manipulations that define the human visual-search
literature, matches the human reaction-time signature. Because effort and accuracy
are partially substitutable currencies for difficulty (Section~\ref{sec:asym}),
the two approaches probe different facets of the same phenomenon.

\section{Methods}
\label{sec:methods}

\begin{figure}[t]
\centering
\includegraphics[width=\textwidth]{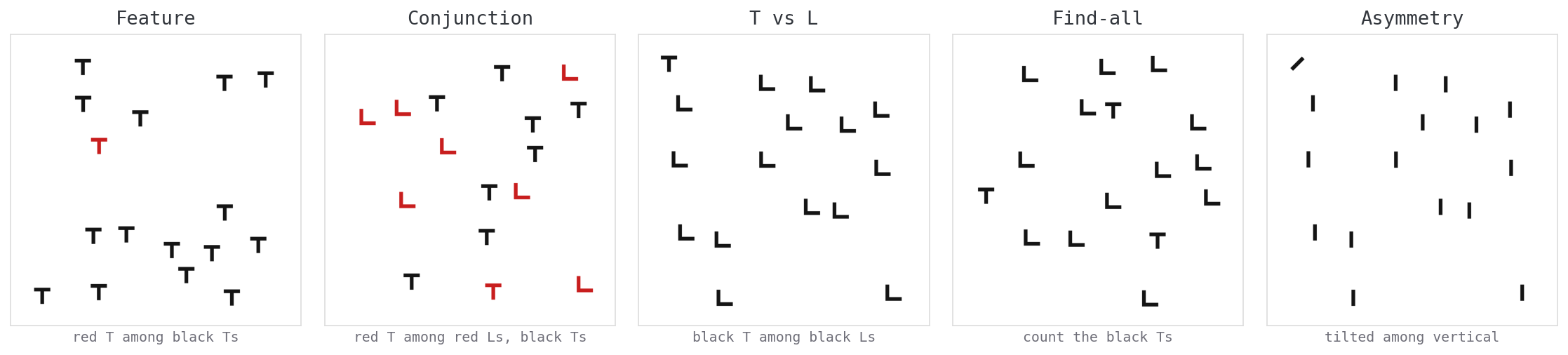}
\caption{Example displays for the five conditions. The first two (feature,
conjunction) retain color; the last three remove it, isolating
spatial-configuration search, enumeration, and a search asymmetry.}
\label{fig:stimuli}
\end{figure}

\paragraph{Stimuli.}
All displays were rendered programmatically as images on a white field
(Figure~\ref{fig:stimuli}). In the feature/conjunction study, items were letters:
feature displays contained a red T among black Ts, and conjunction displays a red
T among red Ls and black Ts, so that neither color nor shape alone identified the
target. I crossed condition (feature, conjunction) with set size
($4,8,16,32$) and target presence, with 25 trials per cell ($400$ displays). The
three follow-up studies removed color. The T-vs-L study placed a single black T
among black Ls (set sizes $4,8,16,32$; present/absent; 25 per cell; $200$
displays). The enumeration study placed one to four black Ts among black Ls (set
sizes $8,16,32$; 25 per cell; $300$ displays). The asymmetry study used short
bars, either a tilted ($45^\circ$) target among vertical distractors or the
reverse (set sizes $4,8,16,32$; present/absent; 25 per cell; $400$ displays).

\paragraph{Models and the effort measure.}
I tested mid-tier and frontier models from two providers. In the
feature/conjunction study these were a mid-tier model in standard and
extended-thinking modes (Claude Sonnet 4.6), a standard multimodal model
(GPT-4o), a mid-tier reasoning model (o4-mini), and two frontier reasoning models
(Claude Opus 4.8 and GPT-5.5). The three follow-up studies used the two frontier
models. Each trial was a single API call carrying the image and one line of text.
For detection the prompt was, e.g., \emph{``Look at this image carefully. Is there
a red letter T in this image? Answer with exactly one word: yes or no.''}; for
enumeration, \emph{``How many letter Ts are in this image? Answer with a single
number.''} The asymmetry task asked either \emph{``Is there a tilted bar?''} or
\emph{``Is there a vertical bar?''} depending on the target, which couples the
visual manipulation to a prompt word---a confound I return to in
Section~\ref{sec:limits}. Standard-mode calls answer in a single pass with no room
to deliberate, the closest analog to brief-exposure human search. Reasoning calls
emit intermediate tokens first. I recorded the response, correctness, and token
usage, and use the number of output/reasoning tokens per trial as the effort
measure. I note one methodological caveat that recurs in the results: Claude Opus
4.8 was run with \emph{adaptive} thinking, in which the model itself decides how
much to deliberate; on tasks it treats as easy it emits essentially no reasoning
tokens, so its effort axis is informative only where it chooses to engage.

\paragraph{Human benchmark.}
For the feature/conjunction comparison I use the public data set of
\citet{wolfe2010}, restricted to its feature and conjunction tasks
($75{,}910$ trials from 9--10 observers). Because human displays remained visible
until response, observers trade time, not accuracy: human accuracy is near
$98\%$ throughout, and the dependent measure is mean correct reaction time. I
compare the \emph{shape} of the human reaction-time function with the shape of the
model effort function, not their absolute scales (milliseconds versus tokens). No
comparable human data set exists for the three follow-up paradigms in these exact
stimuli; for those I compare model behavior against qualitative predictions from
the human literature (and, for enumeration, the subsequent-search-miss pattern
sketched in Figure~\ref{fig:findall}), not against new human measurements.

\paragraph{Analysis.}
For each cell I report mean accuracy and mean effort. I summarize effort
functions by their \emph{slope} across set size, the diagnostic serial signature
(tokens or milliseconds per item), estimated by ordinary least squares on the
trial-level data and reported with $95\%$ confidence intervals; for human reaction
time the slope is fit on the published cell means. To quantify the
human--model correspondence I correlate mean human reaction time with mean model
effort across the sixteen matched cells (two conditions $\times$ four set-size
ranks $\times$ target presence), reporting Pearson $r$ and Spearman $\rho$. Counting
fidelity is assessed by regressing reported on true count and by a Wilcoxon
signed-rank test on the signed error (reported minus true). Accuracy differences
(e.g.\ the asymmetry conditions) are tested by Fisher's exact test and by exact
binomial tests against chance; effort differences between conditions by the
Mann--Whitney $U$ test. Proportions carry Wilson confidence intervals. With 25
trials per cell these analyses are powered to detect the direction and shape of
effects, not fine-grained magnitudes.

\section{Results}

\subsection{Experiment 1: feature versus conjunction search}
\label{sec:fc}

\begin{figure}[t]
\centering
\includegraphics[width=\textwidth]{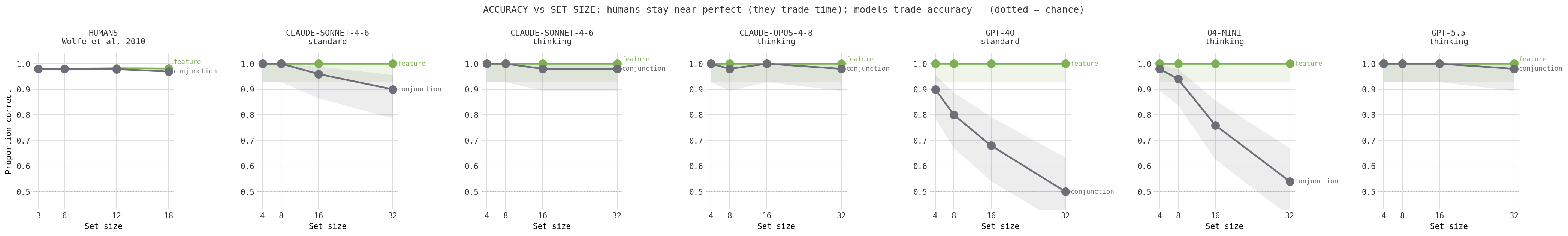}
\caption{Proportion correct by set size for humans and four model configurations.
Feature search (green) is at ceiling for every model; conjunction accuracy (gray)
falls with set size, gently for the mid-tier model with thinking and steeply for
GPT-4o and o4-mini. Humans (leftmost) stay near-perfect in both conditions
because they trade time for accuracy. Shaded bands are $95\%$ Wilson CIs.}
\label{fig:exp1acc}
\end{figure}

\paragraph{Accuracy.}
Every model was perfect on feature search at every set size. Conjunction search
separated the models sharply (Figure~\ref{fig:exp1acc}). At the largest set size
($32$ items) accuracy fell to chance for GPT-4o ($0.50$) and o4-mini ($0.54$),
declined modestly for the standard mid-tier model ($0.90$), and stayed high for
the same model with thinking enabled ($0.98$) and for both frontier models
($0.98$). Errors were overwhelmingly misses rather than false alarms (false-alarm
rates near zero), which is itself human-like: under load, observers tend to quit
and report ``absent'' rather than hallucinate a target. The accuracy collapse of
the weaker models is therefore a capability limit that the frontier has largely
eliminated.

\begin{figure}[t]
\centering
\includegraphics[width=\textwidth]{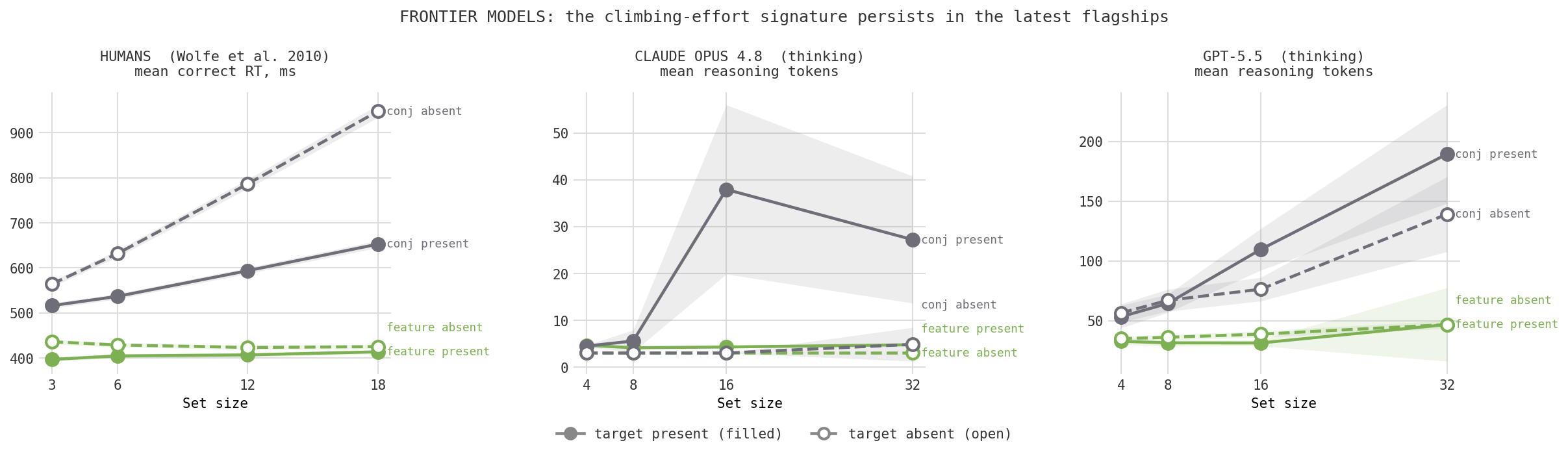}
\caption{Human reaction time and frontier-model reasoning effort by set size.
Feature search is flat and conjunction search climbs in all three panels. The
currency differs (milliseconds for people, reasoning tokens for models) but the
shape is shared. Shaded bands are $95\%$ CIs; Claude Opus 4.8's wide conjunction
band reflects its erratic, near-floor token spending.}
\label{fig:exp1eff}
\end{figure}

\begin{figure}[t]
\centering
\includegraphics[width=\textwidth]{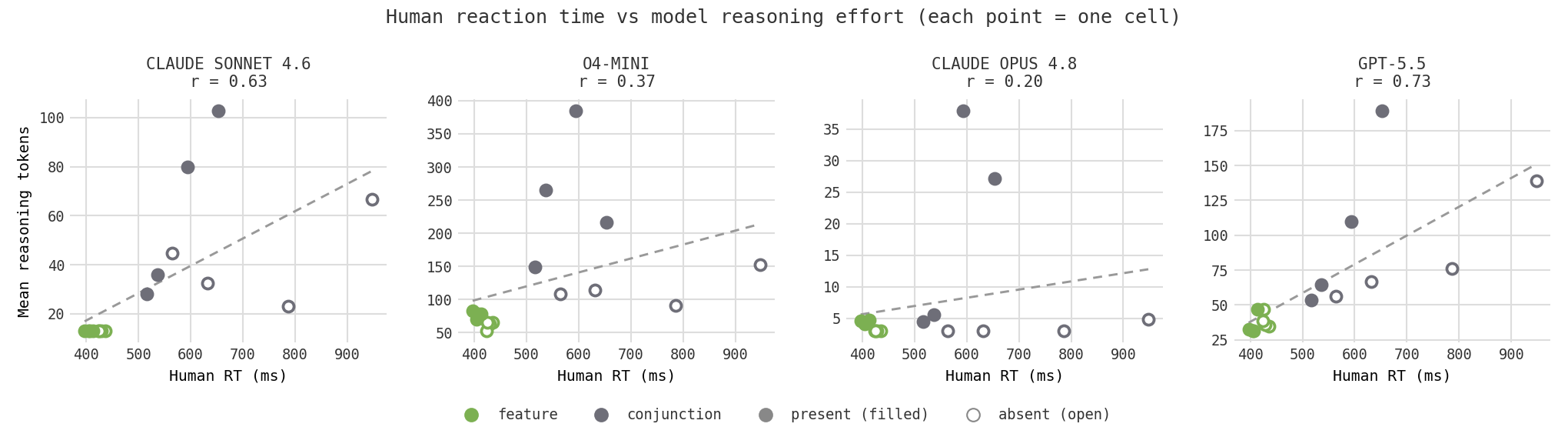}
\caption{Human reaction time versus model reasoning effort across the sixteen
matched cells, per thinking model; each point is one cell (green feature, gray
conjunction; filled present, open absent). The correspondence is strong for
GPT-5.5 and the mid-tier model, modest for o4-mini, and essentially absent for
Claude Opus 4.8, whose effort sits near the floor.}
\label{fig:corr}
\end{figure}

\paragraph{Effort.}
The reasoning models reproduced the human signature in shape
(Figure~\ref{fig:exp1eff}). Feature-search effort was flat across set size for
every model (all feature slopes $\le 0.5$ tokens/item, confidence intervals
including or adjacent to zero). Conjunction-search effort rose with set size, and
significantly so for the models that actually deliberated: on target-present
trials the mid-tier thinking model climbed at $2.76$ tokens/item ($95\%$ CI
$[2.10, 3.41]$, $p<10^{-12}$) and GPT-5.5---most cleanly and monotonically---at
$4.98$ tokens/item ($[3.91, 6.06]$, $p<10^{-14}$). Claude Opus 4.8 also has a
positive slope ($0.90$, $[0.34, 1.46]$, $p=0.002$), but it should be read with
caution rather than as a comparable measurement: Opus spent a median of only about
five tokens per trial, and its curve is non-monotonic (peaking at set size 16,
then falling), reflecting its adaptive thinking rarely engaging
(Section~\ref{sec:limits}). Nobody instructed the models to work harder as
displays grew; they allocated effort the way human reaction time does. Quantifying
the correspondence directly, mean model effort tracked mean human reaction time
across the sixteen matched cells (Figure~\ref{fig:corr}): GPT-5.5 $r=0.73$
($p=0.001$; Spearman $\rho=0.91$) and the mid-tier thinking model $r=0.63$
($p=0.009$; $\rho=0.83$). GPT-5.5 produced the cleanest, most monotonic version of
the human reaction-time curve of any model tested. Two exceptions are informative
(Figure~\ref{fig:corr}): Claude Opus 4.8 correlated only weakly ($r=0.21$, n.s.),
its adaptive thinking leaving effort near the floor (Section~\ref{sec:limits}),
and o4-mini correlated only modestly ($r=0.37$, n.s.) while its conjunction
accuracy fell toward chance (its target-present conjunction slope did not differ
from zero, $[-3.06, 5.50]$)---the profile of giving up rather than searching. The
correlation is computed over aggregated cell means with human and model set sizes
paired by ordinal rank, so it indexes the agreement of the two effort \emph{shapes}
rather than a trial-level fit.

\paragraph{A reversal.}
The human data show a robust ordering: the target-absent conjunction slope is
about $2.8$ times the target-present slope ($25.7$ versus $9.2$ ms/item), the
classic signature of exhaustive search---ruling a target out requires inspecting
more of the display than finding one. The models reversed this. Across thinking
models the target-present conjunction slope \emph{exceeded} the target-absent
slope: $0.90$ ($[0.34, 1.46]$) versus $0.07$ ($[-0.02, 0.15]$) tokens/item for
Claude Opus 4.8, and $4.98$ ($[3.91, 6.06]$) versus $2.94$ ($[2.12, 3.76]$) for
GPT-5.5, the present and absent intervals being non-overlapping or nearly so. The
models spend their effort confirming and localizing a target they have found,
rather than patiently verifying an absence.
Whatever they are doing when they answer ``no,'' it is not the item-by-item sweep
that people perform, and this reversal persisted into the newest models.

\subsection{Resolution control}
\label{sec:blur}

A confound threatens the effort interpretation: a crowded conjunction display
also contains more small letters, so extra tokens might reflect the cost of
resolving shapes rather than the cost of search. To separate these, I held the
high-clutter layouts (set sizes $16$ and $32$) fixed and degraded only the
resolvable detail, by downsampling each image and upsampling it back to the
original size; this blurs the strokes that distinguish a T from an L while
leaving the frame, layout, image size, and color untouched. Feature search rides
along as a control, since color survives blurring.

\begin{figure}[t]
\centering
\includegraphics[width=0.78\textwidth]{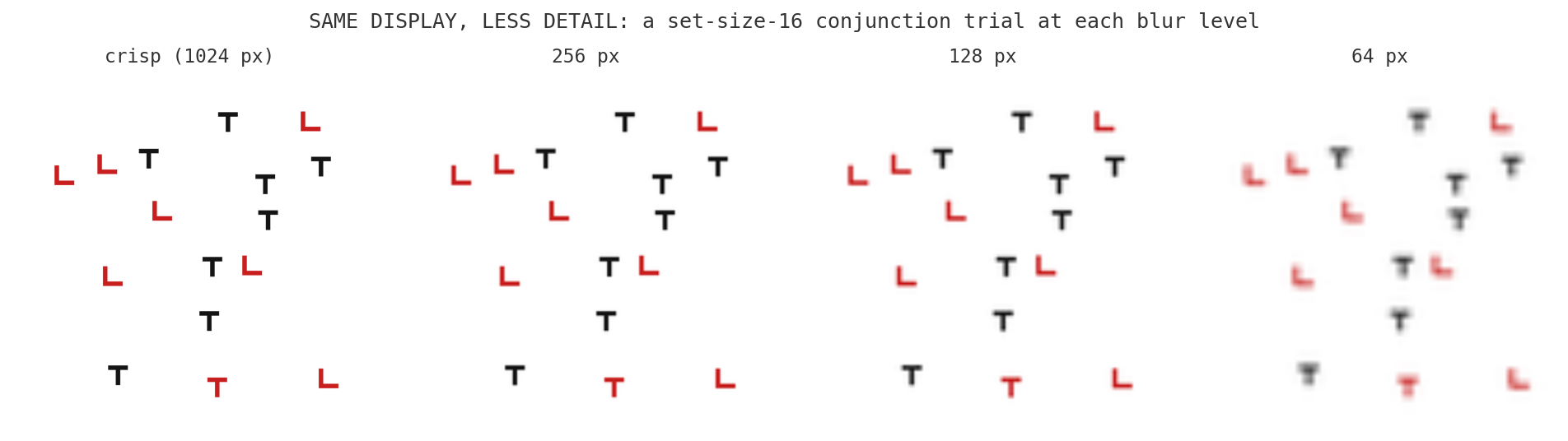}\\[0.4em]
\includegraphics[width=0.86\textwidth]{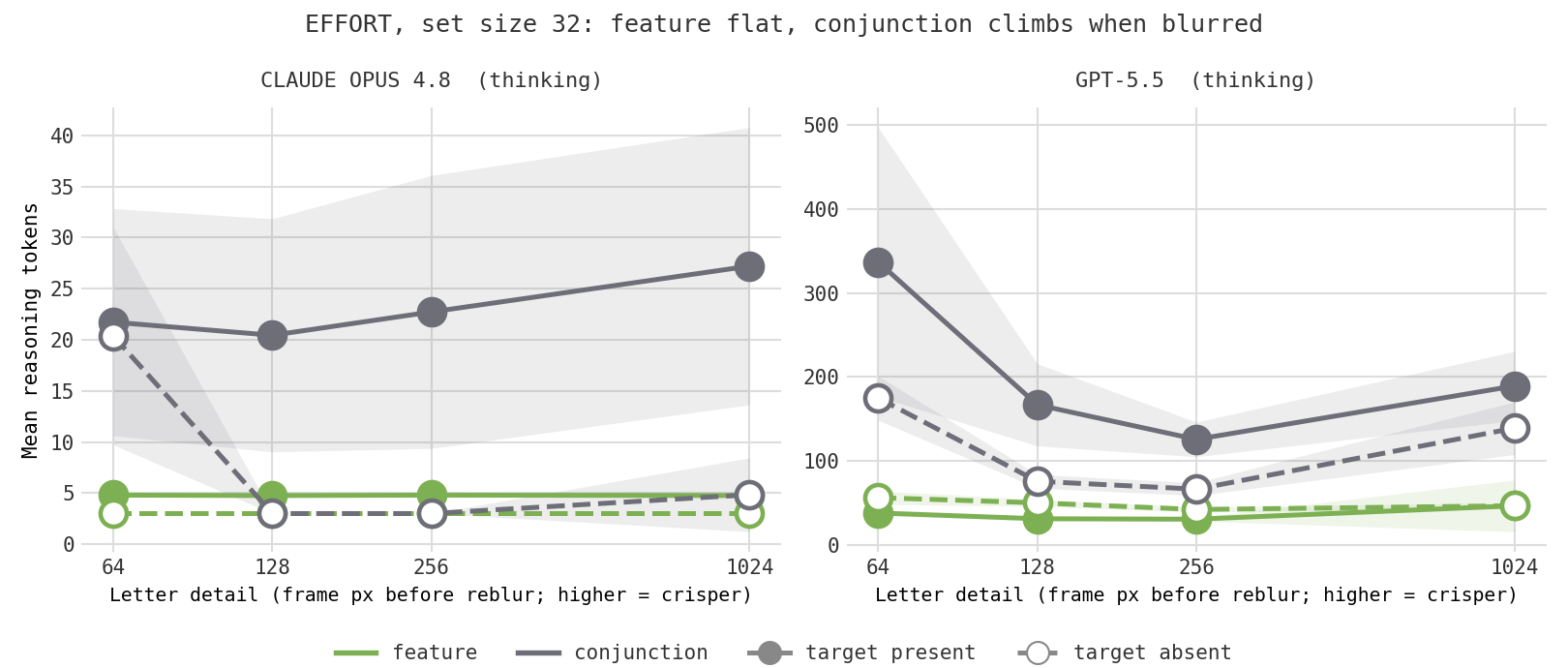}
\caption{Top: one set-size-16 conjunction display at four detail levels, from
crisp (left) to a few-pixel smear (right); color is preserved throughout. Bottom:
reasoning effort at set size 32 across the detail ladder (x-axis: pixels carried
before reblurring; higher is crisper). Feature effort (green) stays flat;
conjunction effort (gray) is higher and rises only at the blurriest end, with the
crisp baseline already costly. Shaded bands are $95\%$ CIs.}
\label{fig:bluref}
\end{figure}

Accuracy never broke: both frontier models found the target at every detail
level, down to letters reduced to a few-pixel smear. Feature-search effort stayed
flat regardless of blur. Conjunction effort showed a perceptual component only at
the extreme: for GPT-5.5 at set size $32$ it traced a shallow U, lowest at
intermediate detail and rising at the blurriest level
(Figure~\ref{fig:bluref}). Crucially, the crisp, easy-to-read baseline was not the
cheap end of the curve; original letters cost as much as or more than mildly
blurred ones. The effort spent on a normal conjunction display is therefore not
mainly the cost of resolving small shapes. It behaves like genuine search, with a
perceptual surcharge that appears only when the image is badly degraded. This
conclusion rests mainly on GPT-5.5: Claude Opus 4.8 spends so few tokens here that
its detail curve is near the floor and too noisy to interpret, so the control
should be read as a single-model result.

\subsection{Experiment 2: spatial-configuration (T-vs-L) search}
\label{sec:tvl}

\begin{figure}[t]
\centering
\includegraphics[width=\textwidth]{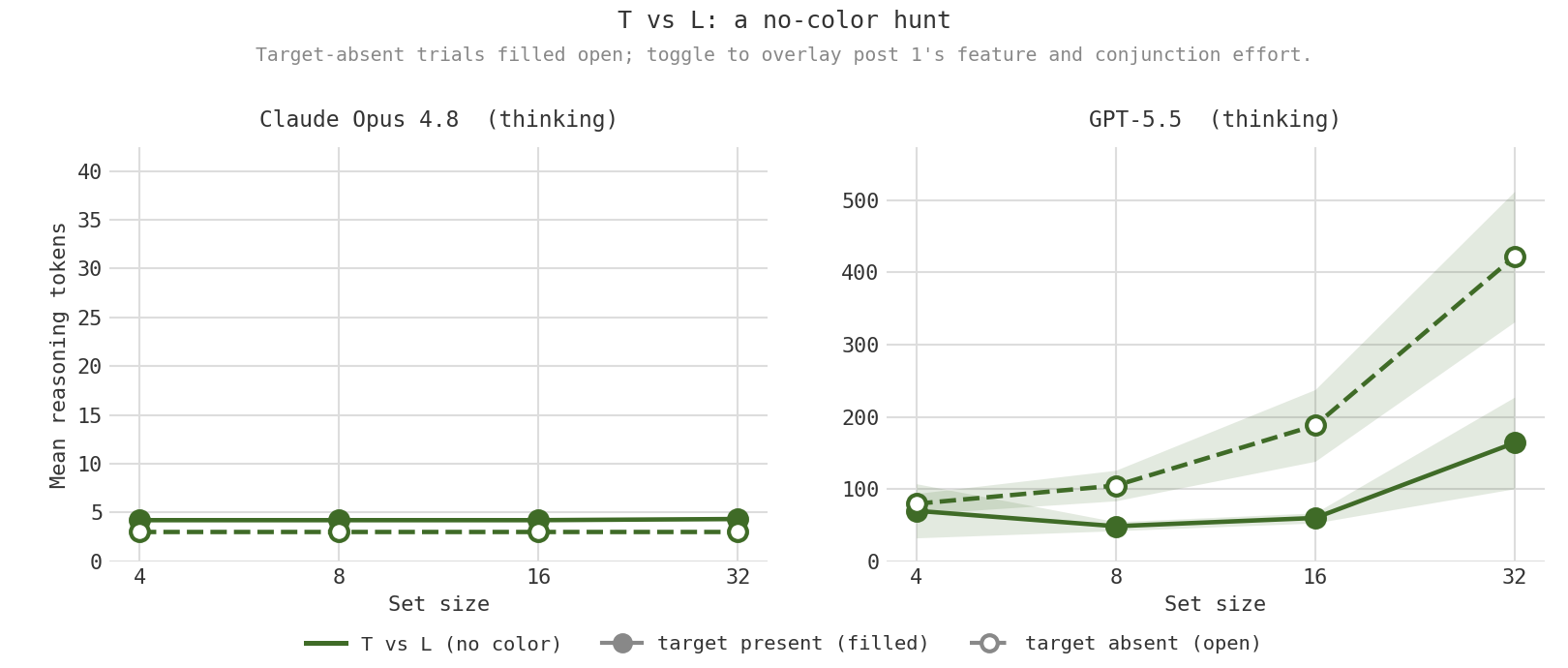}
\caption{T-vs-L reasoning effort by set size, target present (solid) and absent
(open), with $95\%$ CIs. GPT-5.5's effort climbs steeply, most so on absent
trials; Claude remains flat near four tokens. (Overlaying the feature and
conjunction curves from Experiment~1 places GPT-5.5's T-vs-L line on its
conjunction curve and Claude's on its feature floor.)}
\label{fig:tvl}
\end{figure}

Removing color turns detection into a spatial-configuration search: a T and an L
share the same two strokes, so with a single color neither color nor any single
shape feature finds the target, and the only cue is the spatial arrangement of
each letter's parts. Resolving that arrangement requires binding the parts at an
attended location, which classic results predict will be serial and
conjunction-like \citep{wolfe2021}; in humans the difficulty of such search is
captured by a small functional visual field, only a few items processed per
fixation \citep{hulleman2020}. Both models were
near-perfect (Claude correct at every set size; GPT-5.5 slipping only to $0.96$
at $32$ items), so the signal is again in effort (Figure~\ref{fig:tvl}). On
target-absent trials, where search must be most thorough, GPT-5.5's effort rose
from about $80$ tokens at four items to $422$ at thirty-two (slope $12.5$
tokens/item, $[10.0, 15.0]$, $p<10^{-15}$; present-trial slope $3.8$,
$[2.0, 5.5]$), tracking its own conjunction curve from Experiment~1---the serial
signature, now with no color to guide it. Claude, by contrast, spent three to
four tokens per trial at every set size, the same as it spent on the feature
pop-out, and answered correctly anyway: its adaptive thinking judged the task not
worth deliberating over. The same no-color hunt is thus a serial search for one
model and a reflex for the other.

\subsection{Experiment 3: enumeration}
\label{sec:enum}

\begin{figure}[t]
\centering
\includegraphics[width=0.49\textwidth]{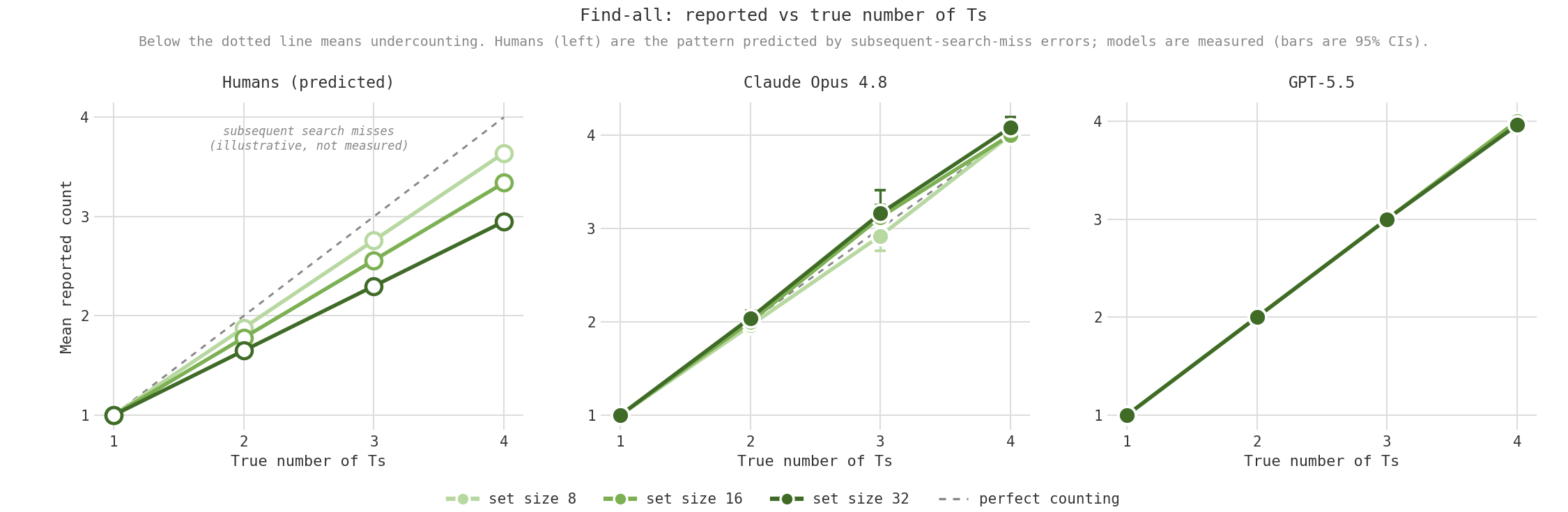}
\hfill
\includegraphics[width=0.49\textwidth]{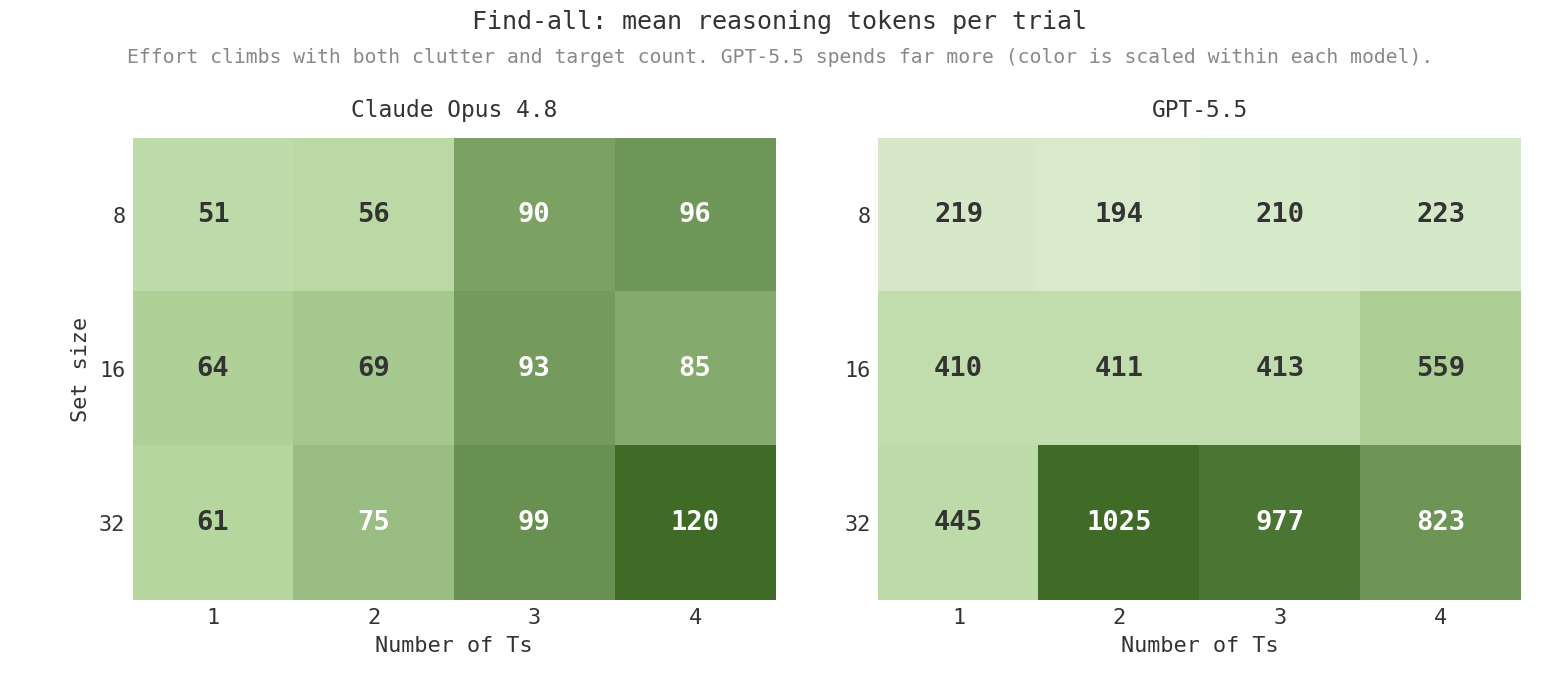}
\caption{Enumeration. Left block: mean reported count versus true count by set
size, with the identity line. The leftmost panel is the pattern predicted for
humans by subsequent-search-miss errors (illustrative, not measured); the two
model panels are measured (bars are $95\%$ CIs) and sit on the identity line.
Right: mean reasoning tokens per trial by set size and target count; effort grows
with both, and GPT-5.5 spends far more (color scaled within each model).}
\label{fig:findall}
\end{figure}

When several targets are present at once, humans exhibit \emph{subsequent search
misses} (SSM; historically ``satisfaction of search''): finding one target makes
them reliably worse at detecting the rest, an error robust across the laboratory,
radiology, and baggage screening and attributed in part to a working-memory cost
of holding the found target \citep{fleck2010,cain2013}. The prediction for my
count-the-Ts task is therefore that a searcher should undercount, missing later
targets as the display fills (Figure~\ref{fig:findall}, left panel, sketches this
predicted pattern; I collected no human data for this task).

The models showed the opposite. Asked to count one to four black Ts among black
Ls, both tracked the true count almost perfectly: exact-count accuracy was $0.967$
for Claude ($95\%$ CI $[0.940, 0.982]$) and $0.993$ for GPT-5.5 ($[0.976, 0.998]$)
overall, and reported count rose with true count at essentially unit slope (Claude
$1.015$, $r=0.97$; GPT-5.5 $0.992$, $r=0.997$), sitting on the identity line rather
than bending below it (Figure~\ref{fig:findall}, center and right). Directly
testing the undercount prediction, the signed error (reported minus true count)
was indistinguishable from zero for both models (Wilcoxon $p=0.13$ for Claude,
$p=0.16$ for GPT-5.5) and did not decline with target count or set size, so there
is no statistical sign of subsequent-search-miss undercounting; if anything
Claude's error trended slightly positive in the most crowded displays.
What the count cost the models was effort, not accuracy
(Figure~\ref{fig:findall}, far right): mean reasoning tokens rose with both target
count and set size, and for GPT-5.5 ran past a thousand tokens per trial on the
most crowded displays---roughly an order of magnitude more than Claude. The models
thus depart from the human enumeration profile: rather than dropping additional
targets under load, they preserve accuracy by spending more compute.

\subsection{Experiment 4: search asymmetry}
\label{sec:asym}

\begin{figure}[t]
\centering
\includegraphics[width=0.49\textwidth]{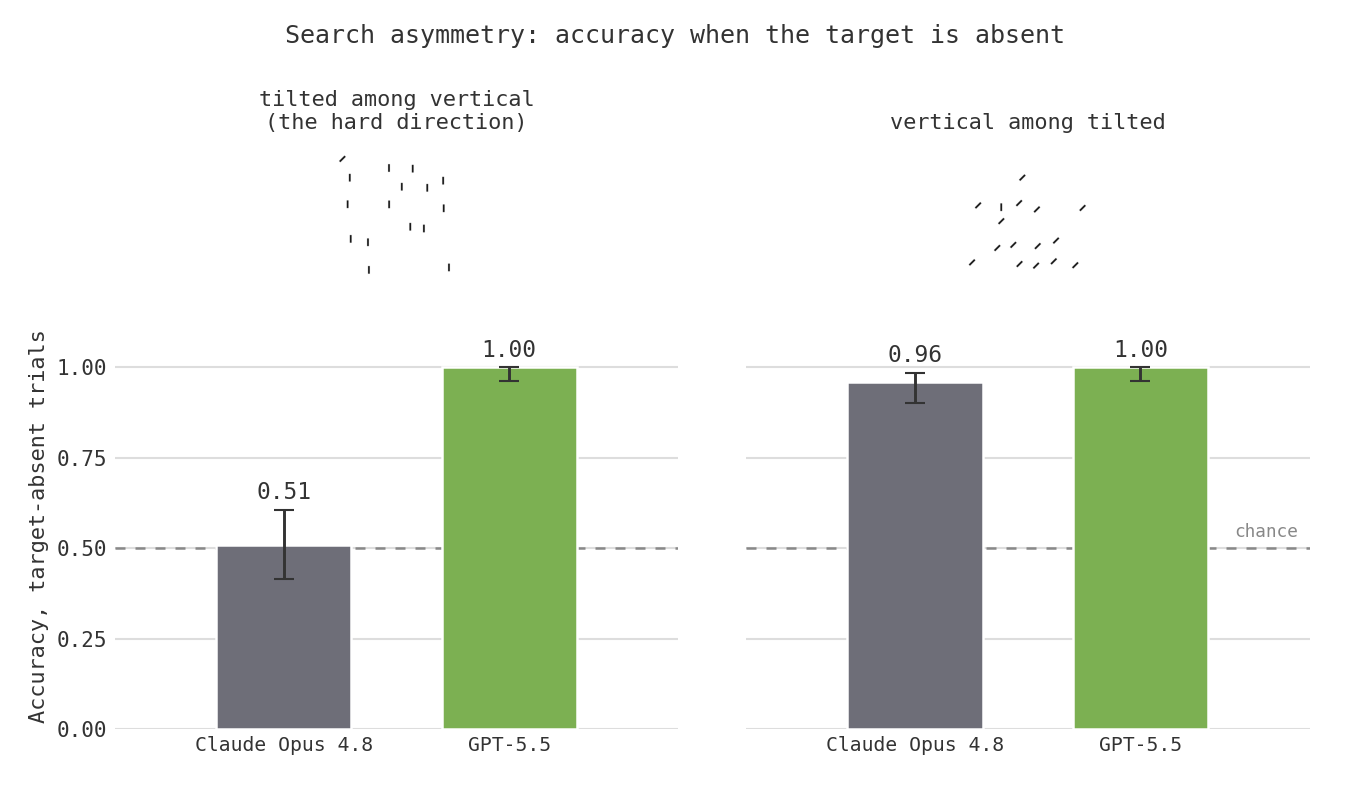}
\hfill
\includegraphics[width=0.49\textwidth]{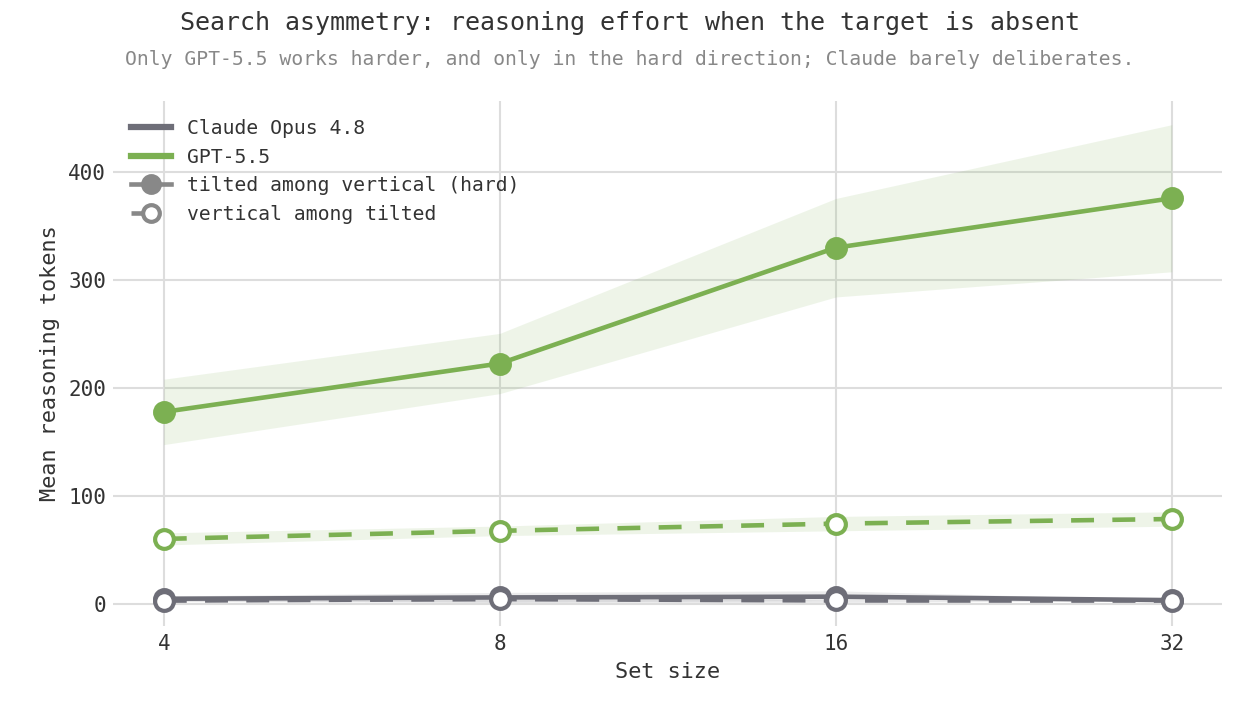}
\caption{Search asymmetry on target-absent trials, both models per panel. Left:
accuracy (bars are $95\%$ Wilson CIs); only Claude breaks, falling to chance when
ruling out a tilted bar among vertical ones. Right: reasoning tokens by set size
with $95\%$ CIs; only GPT-5.5 works harder, and only in that same hard direction.}
\label{fig:asym}
\end{figure}

The asymmetry task is the strongest test because its two directions use identical
elements. In humans, a tilted bar among vertical distractors is found efficiently
while the reverse is inefficient \citep{treisman1988}. Both models showed an
asymmetry, both localized it to the same hard case---the target-absent
tilted-among-vertical display, in which the model must confirm that no tilted bar
is present among uniformly vertical bars---and they expressed it in different
currencies (Figure~\ref{fig:asym}). GPT-5.5 kept accuracy perfect everywhere
($100/100$ in both directions) but spent far more reasoning to clear the hard
direction (median $241$ vs $69$ tokens; Mann--Whitney $p<10^{-30}$), with effort
rising in set size only for that direction (slope $7.0$ tokens/item,
$[4.8, 9.1]$, $p<10^{-8}$). Claude, which again barely deliberated, paid in errors
instead: its accuracy on the hard direction's absent trials was $51/100$,
indistinguishable from chance (exact binomial $p=0.92$), against $96/100$ for the
mirror direction (Fisher's exact, $p<10^{-13}$); on those trials it reports a
tilted bar that is not there. A single underlying asymmetry thus surfaces as an
effort gradient in one model and as an accuracy cliff in the other.

Notably, the models' shared hard case is not the human one. For people a tilted
target pops out, so confirming its absence should be easy; the models instead find
exactly that confirmation the most demanding part of the task. They are lopsided,
as humans are, but lopsided in a different direction.

\begin{table}[t]
\centering
\small
\caption{Summary of signatures. ``Human-like'' indicates the model pattern matches
the direction of the human effect; effort refers to reasoning tokens.}
\label{tab:summary}
\begin{tabular}{@{}lll@{}}
\toprule
Phenomenon & Human signature & VLM finding \\
\midrule
Feature search & flat RT slope & flat effort (all models) \\
Conjunction search & rising RT slope & rising effort (thinking models) \\
Conjunction accuracy & near-ceiling (time trade) & collapses for mid-tier, holds at frontier \\
Absent vs present slope & absent steeper ($\sim$2.8:1) & present steeper (reversed) \\
T-vs-L (no color) & serial, rising slope & rising effort (GPT-5.5); flat (Claude) \\
Enumeration $>4$ & accuracy drops (counting) & accuracy held, effort rises \\
Asymmetry & tilted-among-vertical easy & hardest case is confirming its absence \\
\bottomrule
\end{tabular}
\end{table}

\section{Discussion}

Read as a whole (Table~\ref{tab:summary}), the results support a measured
conclusion: frontier VLMs reproduce the coarse behavioral signatures of human
visual search, and the way they fail to reproduce the finer ones is itself
diagnostic.

\paragraph{What matches.}
The flat-versus-sloped contrast that defines the literature appears cleanly in
reasoning-token effort. Feature search is free; conjunction search and the
no-color T-vs-L hunt cost effort that grows with set size; the growth is
monotonic for the strongest model. That a within-model effort signal, never
optimized to mimic reaction time, traces the human curve suggests that the
allocation of internal computation in these models is organized along a
difficulty axis that aligns with human attentional demand. This dovetails with
\citet{budny2025}, who reach a related conclusion from the opposite direction,
using human reaction time to predict where VLM accuracy falls.

\paragraph{What diverges, and why it is interesting.}
Three divergences stand out. First, the present/absent reversal: humans work
harder to confirm absence, whereas the models work harder to confirm presence.
The natural reading is that the models are not running the patient, exhaustive
sweep that the human absent-trial slope reflects. A more mundane alternative is
also consistent with the data: the models may adopt a \emph{find-one-and-justify}
strategy, spending reasoning to confirm and localize a target that is present and
terminating early when they decide to answer ``no.'' That is a difference in
stopping policy rather than proof that no serial process occurs; either way, the
ordering is not the human one, and adjudicating between the two accounts would
require trial-level reasoning traces. Second, enumeration: the models hold accuracy
where humans would lose count, paying in compute rather than errors, because they
can grind through a display rather than apprehend it at a glance. Third, and most
striking, the divergence \emph{between} models. Because Claude Opus 4.8 allocates
deliberation adaptively, it treats detection as trivial and answers without
visible reasoning, so the same search that GPT-5.5 expresses as a graded effort
gradient surfaces in Claude only when it fails---as the accuracy cliff in the
asymmetry task. Effort and accuracy are partially substitutable currencies for
difficulty, and which one a model spends is a property of the model, not of the
task.

\paragraph{Reasoning tokens as a measure.}
The reversal and the adaptive-thinking results also delimit what the effort
measure can and cannot show. Token count is a faithful index of how much a model
chooses to deliberate, and its shape across psychophysical manipulations is
informative. But a model that declines to deliberate is invisible on this axis
even when it is performing the task, and a model's token budget reflects training
and decoding policy as much as perceptual difficulty. The measure is therefore
best read alongside accuracy, as I do in Experiments~3 and~4, rather than on its
own.

\paragraph{Limitations.}
\label{sec:limits}
Several limitations bound these conclusions, and each suggests a concrete
follow-up.

\emph{Validity of the effort proxy.} Reasoning-token count is an analog of
reaction time, not a substitute. It is shaped by decoding policy, output verbosity
norms, and training as much as by perceptual difficulty, so its absolute scale is
not interpretable across models. My claims therefore rest on the \emph{slope}
across set size---the established serial signature
\citep{treisman1980,trick1994}---rather than on token magnitudes, and I report
those slopes with confidence intervals. I cannot fully exclude that part of the
slope reflects a model's tendency to narrate longer about busier images;
relating token counts to wall-clock latency, and to chain-of-thought content, is
needed to tighten the mapping.

\emph{The adaptive-thinking confound.} Because Claude Opus 4.8 chooses its own
deliberation, its effort axis is undefined on tasks it treats as trivial; its flat
functions and weak RT correlation ($r=0.21$) mean ``declined to engage,'' not
``searched cheaply.'' Effort is thus not on equal footing across models. The fix
is to re-run with matched, externally fixed thinking budgets, and to pair every
effort measure with an accuracy measure, as I do in Experiments~3 and~4.

\emph{Causal versus observational design.} Most analyses are observational. The
one manipulation I exploit---removing the color cue between the conjunction study
and the T-vs-L study---is suggestive of a causal dissociation but is not clean,
because the two displays also differ in distractor composition. A within-subjects
version that toggles only the guiding color on otherwise identical displays, in the
spirit of the color-marking manipulation of \citet{budny2025}, would isolate the
contribution of feature guidance.

\emph{No internal analysis.} I treat the models as black boxes, as I must for
closed frontier systems. Attention-level analysis on open VLMs, of the kind used
by \citet{campbell2024} and \citet{budny2025} to localize failures to object
individuation, would connect my behavioral signatures to mechanism.

\emph{Residual confounds.} Two are worth naming. In the asymmetry task the two
directions differ not only in their visual roles (tilted vs vertical target) but
in the prompt word (``Is there a tilted bar?'' vs ``\dots vertical bar?''). Given
the thesis that VLMs lean on linguistic grounding, the word itself could
contribute to the effort difference; a counterbalanced-prompt control would
separate the visual asymmetry from a lexical one. Separately, I report many
significance tests without correction for multiple comparisons; the headline
effects are far below any reasonable corrected threshold, but the smaller
$p$-values should be read as descriptive.

\emph{Scope and the human side.} The design is small-$n$ and deep (25 trials per
cell), spans few models, and uses synthetic letters and bars rather than natural
scenes, so generalization to richer images and to other model families is
untested. I also do not collect new human data; a symmetric control---restricting
human exposure time, which is known to abolish serial binding and exact counting
\citep{treisman1980,trick1994}---would test the prediction that humans, denied time
to search, collapse toward the model profile.

\paragraph{Conclusion.}
The tools built to study human attention transfer to machine vision with little
adaptation and at low cost, and they discriminate sharply among models. Frontier
VLMs carry the fingerprint of human visual search in the gross shape of their
effort and accuracy, while departing from it in the details of absence, counting,
and how each model spends difficulty. The places where the analogy breaks are
where the interesting questions about machine perception begin.

\section*{Acknowledgements}
I am grateful to Jeremy M. Wolfe for suggesting the three follow-up experiments
(T-vs-L, enumeration, and the search asymmetry) after reading an early write-up of
the feature/conjunction study, and for the public human benchmark data from his
laboratory on which the comparisons rest. The framing and any errors are my own.

\paragraph{Data and code availability.}
Stimulus generators, model-querying code, raw results, and analysis scripts are
available on request.

{\setlength{\parskip}{0pt}

}

\end{document}